\documentclass[sigconf]{acmart}
\usepackage{enumitem}
\usepackage{multirow}
\usepackage{booktabs}
\usepackage{hyphenat}
\usepackage[ruled,vlined,linesnumbered]{algorithm2e}
\usepackage{xcolor}
\usepackage{amsmath}
\usepackage{tikz}
\usepackage{ulem}
\usepackage{balance}
\usepackage{placeins}
\usetikzlibrary{arrows.meta,positioning,shapes,fit}
\usepackage{tcolorbox}
\tcbuselibrary{listings, breakable, skins}
\newtcblisting{promptbox}[1]{
  colback=gray!5,
  colframe=gray!40,
  colbacktitle=gray!55,
  coltitle=white,
  fonttitle=\bfseries\small,
  title={#1},
  breakable,
  enhanced,
  boxrule=0.3pt,
  listing only,
  listing options={
    basicstyle=\footnotesize\ttfamily,
    breaklines=true,
    breakindent=0pt,
    columns=fullflexible,
    keepspaces=true,
    upquote=true,
    showstringspaces=false,
    aboveskip=0pt,
    belowskip=0pt,
    xleftmargin=0pt,
    xrightmargin=0pt,
  },
  left=2pt, right=2pt, top=1pt, bottom=1pt,
  toptitle=1pt, bottomtitle=1pt,
  middle=0pt,
}

\setlist[itemize]{leftmargin=*, labelsep=0.4em, align=left}

\AtBeginDocument{%
}

\copyrightyear{2026}
\acmYear{2026}
\setcopyright{cc}
\setcctype{by}
\acmConference[KDD '26]{Proceedings of the 32nd ACM SIGKDD Conference on Knowledge Discovery and Data Mining V.2}{August 09--13, 2026}{Jeju Island, Republic of Korea}
\acmBooktitle{Proceedings of the 32nd ACM SIGKDD Conference on Knowledge Discovery and Data Mining V.2 (KDD '26), August 09--13, 2026, Jeju Island, Republic of Korea}
\acmDOI{10.1145/3770855.3818931}
\acmISBN{979-8-4007-2259-2/2026/08}

\begin{document}

\title{ClinicalAgents: Multi-Agent Orchestration for Clinical Decision Making with Dual-Memory}

\author{Zhuohan Ge}
\affiliation{%
  \institution{The Hong Kong Polytechnic University}
  \city{Hong Kong SAR}
  \country{China}
}
\email{zhuohan.ge@connect.polyu.hk}

\author{Haoyang Li}
\authornote{Corresponding author.}
\affiliation{%
  \institution{The Hong Kong Polytechnic University}
  \city{Hong Kong SAR}
  \country{China}
}
\email{haoyang-comp.li@polyu.edu.hk}

\author{Yubo Wang}
\affiliation{%
  \institution{The Hong Kong University of Science and Technology}
  \city{Hong Kong SAR}
  \country{China}
}
\email{ywangnx@connect.ust.hk}

\author{Nicole Hu}
\affiliation{%
  \institution{The Hong Kong Polytechnic University}
  \city{Hong Kong SAR}
  \country{China}
}
\email{hulan.hu@connect.polyu.hk}

\author{Chen Jason Zhang}
\affiliation{%
  \institution{The Hong Kong Polytechnic University}
  \city{Hong Kong SAR}
  \country{China}
}
\email{jason-c.zhang@polyu.edu.hk}

\author{Qing Li}
\affiliation{%
  \institution{The Hong Kong Polytechnic University}
  \city{Hong Kong SAR}
  \country{China}
}
\email{csqli@comp.polyu.edu.hk}

% \author{
% 	Zhuohan Ge{$^{1}$}, Haoyang Li{$^{1}$}, Yubo Wang{$^2$}, Nicole Hu{$^1$}, Chen Jason Zhang{$^1$}, Qing Li{$^1$}
% }
% \affiliation{
% 	\institution{$^1$The Hong Kong Polytechnic University, Hong Kong SAR}
% 	\city{$^2$The Hong Kong University of Science and Technology, Hong Kong SAR}
% 	\country{}
% }

\renewcommand{\shortauthors}{Zhuohan Ge et al.}

\begin{abstract}
While Large Language Models (LLMs) have demonstrated potential in healthcare, they often struggle with the complex, non-linear reasoning required for accurate clinical diagnosis. 
Existing methods typically rely on static, linear mappings from symptoms to diagnoses, failing to capture the iterative, hypothesis-driven reasoning inherent in human clinicians. 
To bridge this gap, we introduce ClinicalAgents, a novel multi-agent framework designed to simulate the cognitive workflow of expert clinicians. 
Unlike rigid sequential chains, ClinicalAgents employs a dynamic orchestration mechanism modeled as a Monte Carlo Tree Search (MCTS) process. 
This allows an orchestrator to iteratively generate hypotheses, actively verify evidence, and trigger backtracking when critical information is missing. 
The foundation of this framework is a Dual-Memory architecture: a mutable working memory that maintains the evolving patient state for context-aware reasoning, and a static experience memory that retrieves clinical guidelines and historical cases via an active feedback loop. 
Extensive experiments demonstrate that ClinicalAgents achieves the best performance among evaluated baselines, significantly enhancing both diagnostic accuracy and explainability compared to strong single-agent and multi-agent baselines.
Our code is released at \url{https://github.com/ZhuohanGe/ClinicalAgents-Code}.
\end{abstract}




\maketitle

\section{Introduction}
Clinical decision-making (CDM) is a complex, continuous process that involves evaluating evidence and patient symptoms to provide accurate diagnoses and effective treatment~\cite{elstein2002clinical}. 
CDM is one of the most crucial and challenging aspects of clinical medicine and a core responsibility of physicians. 
It requires physicians to continuously collect information based on patients' symptoms in a highly uncertain environment and comprehensively process multimodal data, including electronic health records (EHRs), medical imaging, and laboratory tests, dynamically update differential diagnoses, and align with medical guidelines~\cite{tiffen2014enhancing,sutton2020overview}.
To cope with such complexity, Artificial Intelligence (AI) has emerged as a powerful auxiliary tool in healthcare~\cite{yu2018artificial,guidance2021ethics}.
By automating the analysis of massive, diverse medical data, AI significantly reduces physicians' workload and enhances the overall efficiency of the decision-making process.

Modern AI approaches in healthcare can be broadly categorized into two paradigms: deep learning (DL) and pre-trained language models (PLMs). 
DL models, such as UNETR~\cite{hatamizadeh2022unetr} and HiTANet~\cite{luo2020hitanet}, have demonstrated strong capability in analyzing structured data and medical imaging. 
With the introduction of transformer-based PLMs like Med-BERT \cite{rasmy2021med} and Hi-BEHRT \cite{li2022hi}, AI systems have further advanced by capturing rich semantic patterns from structured clinical records. 
However, both DL and PLM-based models are fundamentally limited in their ability to perform the complex, multi-step reasoning and iterative hypothesis evaluation required for clinical decision-making. 
They typically lack the capacity to dynamically align with clinical guidelines, adapt to new evidence during the diagnostic process, and provide robust rationales that are essential for accuracy and trustworthiness in clinical practice.

In recent years, LLMs have gained widespread attention for their superior contextual understanding and reasoning abilities~\cite{mesinovic2025explainability}.
Early research aimed to improve clinical reasoning by enhancing individual LLMs with advanced prompting strategies \cite{singhal2025toward, dhuliawala2024chain} and reinforcement learning \cite{zhang2023huatuogpt, yu2025finemedlm}. However, as demonstrated by recent studies \cite{singhal2025toward, dhuliawala2024chain, zhang2023huatuogpt, yu2025finemedlm, ge2025survey}, single LLMs remain insufficient for complex clinical decision-making processes, motivating system-level designs that integrate retrieval, verification, and specialized modules. 
In response, Multi-Agent Systems (MAS) have been introduced, where multiple specialized agents collaborate to solve tasks beyond the capacity of a single model. This approach more accurately reflects the multidisciplinary nature of clinical practice.
In MAS, memory management is a key design component that maintains a shared, persistent record of evidence and intermediate conclusions across agents, enabling consistent coordination across multi-round interactions.
Building on this paradigm, recent frameworks employ role-playing agents (e.g., specialists and critics)~\cite{tang2024medagents,wang2025colacare} and incorporate debate mechanisms~\cite{kaesberg2025voting,chen2024reconcile} or realistic workflow simulations~\cite{li2024agent,liu2024medchain} to improve reasoning consensus and adaptability, where agent state is commonly maintained via conversation history and intermediate summaries.

However, existing MAS for CDM still face significant challenges.
First, many existing MAS are implemented with a static, pre-scripted interaction workflow, where agents act in a predetermined order, and the control flow provides limited support for replanning or backtracking when new evidence emerges.
Such rigidity undermines flexibility and error recovery, making it difficult to accommodate the dynamic and uncertain nature of CDM.
Second, existing MAS often treat memory as a window-bounded dialogue context, which can easily fragment, causing agents to lose critical details and drift from the overall objective.
In CDM, this information loss and fragmentation can lead to unstable information exchange and inconsistent intermediate states across agents, ultimately resulting in non-convergent diagnoses.
Third, many existing LLM-based diagnostic studies still treat diagnosis as a mostly static generation task, where a model reads a fixed clinical case and outputs diagnostic results.
In contrast, clinical diagnosis is often described as a hypothetico-deductive process~\cite{patel1986knowledge}: clinicians start with a set of hypotheses, collect evidence to narrow the search space, and iteratively revise the hypotheses until they reach diagnostic closure~\cite{duong2023scoping,higgs2024clinical}.
Without this method of reasoning, MAS may overestimate how sufficient the given information is and settle on a diagnosis even when key evidence or details are missing, increasing the risk of misdiagnosis~\cite{zhou2025uncertainty,norman2009iterative}.

To bridge these fundamental gaps, we present ClinicalAgents, a novel multi-agent framework that operationalizes the expert clinician's cognitive workflow. 
Unlike existing MAS that rely on static, sequential workflows, ClinicalAgents implements a dynamic, hypothesis-driven reasoning mechanism. 
This mechanism orchestrates specialized agents to iteratively generate diagnostic hypotheses, actively seek verifying evidence, and refine diagnoses through continuous falsification cycles. 
To maintain a coherent state across multi-round interactions, we introduce a dual-memory design: a structured working memory that preserves the evolving patient state for context-aware revision, and an experience memory that encodes reusable clinical knowledge.
Finally, we introduce a backtracking mechanism formulated as a Monte Carlo Tree Search (MCTS) based search process~\cite{silver2016mastering,browne2012survey}, ensuring that clinical decisions are both rigorous and transparent.
In summary, our paper makes the following major contributions:
\begin{itemize}
    \item We propose ClinicalAgents, a CDM-oriented multi-agent framework that explicitly instantiates a hypothesis-guided, evidence-seeking diagnostic loop aligned with clinicians’ hypothetico-deductive reasoning.
    \item We design an orchestration mechanism that adaptively selects, reorders, and iterates agent actions through MCTS, overcoming the rigidity of pre-scripted workflows and enabling evidence-triggered backtracking.
    \item We integrate a dual-memory architecture that combines a mutable working memory for state tracking with an experience memory, and uses a hybrid RAG mechanism that supports both passive augmentation and active evidence-seeking, reducing information loss and redundancy, and improving reasoning consistency across multi-round interactions.
    \item Our experiments demonstrate that ClinicalAgents achieves state-of-the-art performance on a multi-stage clinical workflow benchmark compared with existing single-agent and multi-agent baselines. 
    Furthermore, ablation studies, backbone sensitivity analyses, and detailed case studies further validate the framework's robustness.
\end{itemize}
\section{Related Work} \label{sec:related_work}
\subsection{LLMs in Clinical Decision-Making (CDM)}
Motivated by the strong performance on medical licensing exams, early works primarily focused on enhancing the reasoning capabilities of single LLMs for static, offline tasks through advanced prompting and fine-tuning strategies. 
For instance, Med-PaLM 2~\cite{singhal2025toward} introduced ensemble refinement, a strategy that iteratively critiques and refines reasoning chains to improve diagnostic accuracy on USMLE-style benchmarks. 
Similarly, HuaTuoGPT~\cite{zhang2023huatuogpt} leverages hybrid feedback from distilled data and physicians to align models with medical protocols, while FineMedLM-o1~\cite{yu2025finemedlm} incorporates test-time training to enhance reasoning stability during inference. 
While these methods excel in single-turn QA, they fundamentally treat CDM as a static prediction task rather than a dynamic process.

To address the issues of hallucinations and insufficient theoretical depth in long-chain reasoning, researchers have introduced verification and retrieval mechanisms. 
Chain-of-Verification (CoV)~\cite{dhuliawala2024chain} introduces explicit verification steps to detect and correct factual errors, thereby mitigating hallucination risks. 
Recognizing that CDM is not a single QA session, recent research has begun exploring agentic scaffolds to support multi-step reasoning. 
ReAct~\cite{yao2022react} interleaves reasoning traces with tool-using actions (e.g., retrieval), providing a general paradigm that motivates the shift from passive LLMs to active agentic systems. 
This approach attempts to bridge the gap between static knowledge and dynamic decision-making.
However, these lines of work still largely model CDM as offline reasoning over a fixed case, lacking an explicit mechanism to surface evidence and drive iterative evidence acquisition.

\subsection{Multi-Agent Systems in CDM}
To simulate the multidisciplinary nature of clinical practice, research has increasingly pivoted from monolithic models to MAS. 
Foundational frameworks like MedAgents~\cite{tang2024medagents} utilize role-playing mechanisms where specialized agents (e.g., diagnosticians, pharmacologists) collaborate to synthesize comprehensive medical reports. 
This approach is extended by ColaCare~\cite{wang2025colacare}, which bridges EHR modeling with LLM reasoning through collaboration to enhance interpretability. 
To improve adaptability, MDAgents~\cite{kim2024mdagents} dynamically adjusts team size based on medical complexity, routing simple cases to solo agents and complex ones to groups.

Beyond simple collaboration, improving consensus and learning capabilities is a key focus. 
Kaesberg et al.~\cite{kaesberg2025voting} and ReConcile~\cite{chen2024reconcile} demonstrated that while voting is efficient, debate mechanisms are superior for reasoning tasks. 
AMIE~\cite{tu2025towards} highlights the potential of conversational agents optimized through self-play, while Agent Hospital~\cite{li2024agent} introduces simulation-based learning, allowing agents to evolve through interactions in a virtual environment.
Tree-of-Reasoning~\cite{peng2025tree} further structures these interactions into interpretable evidence trees.
Recent efforts have aimed to model realistic clinical processes. 
MedChain~\cite{liu2024medchain} provides a highly realistic end-to-end diagnostic workflow, covering five key stages with personalization and sequential decision dependencies. 
Nevertheless, most MAS for CDM still rely on pre-scripted interaction protocols and window-bounded memory, offering limited replanning and backtracking under incomplete or evolving patient information.

\section{Methodology} \label{sec:methods}
In this section, we present an overview of \textbf{ClinicalAgents}, a novel multi-agent system designed to enhance CDM through effective orchestration of specialized agents. 
% In the following, we describe the design and implementation of key components.
\begin{table}[t]
\centering
\small
\caption{Important Notations.}
\label{tab:notation_clean_hr}
\renewcommand{\arraystretch}{0.86}
\begin{tabular}{p{0.2\linewidth} p{0.72\linewidth}}
\toprule
\textbf{Notation} & \textbf{Definition} \\
\midrule
$\mathcal{M}_\text{work}^t$ & Working Memory at step $t$, defined as $\langle\mathcal{E}_t,\mathcal{H}_t,\tau_t\rangle$. \\
$\mathcal{M}_\text{exp}$ & Experience Memory, defined as $\langle \mathcal{D}_{\mathrm{guide}}, \mathcal{D}_{\mathrm{cdc}} \rangle$. \\
$\mathcal{D}_\text{guide}$ & Structured guideline database. \\
$\mathcal{D}_\text{cdc}$ & Historical case database. \\
$\mathcal{K}_t$ & Retrieved guideline knowledge at step $t$. \\
\midrule
$\mathcal{E}_t$ & Accumulated evidence set at step $t$. \\
$\mathcal{H}_t$ & Diagnostic hypothesis set at step $t$. \\
$\mathcal{E}_t^m$ & Missing critical evidence verified by the orchestrator. \\
$\mathcal{E}_t^p$ & Potentially missing evidence suggested by $\mathcal{M}_\text{exp}$. \\
$c_t$ & Confidence of the top hypothesis at step $t$. \\
\midrule
$\mathcal{A}$ & Action space, $\mathcal{A}=\mathcal{A}_\text{agent}\cup\{a_\text{rag},a_\text{back},a_\text{term}\}$. \\
$\tau_t$ & Action trajectory $(a_0,a_1,\dots,a_t)$. \\
$\mathcal{R}_t$ & Reward at step $t$. \\
$\mathcal{T}$ & Terminal state. \\
$\eta$ & Maximum orchestration step. \\
% $N$ & Number of MCTS rollouts per candidate. \\
$Q(\cdot)$ & MCTS action-value estimate. \\
$\lambda$ & Balance factor in PUCT. \\
$\alpha,\beta,\gamma$ & Weighting coefficients in $\mathcal{R}_t$. \\
$\gamma_d$ & Discount factor for cumulative return. \\
\midrule
$\mathcal{O}(\cdot)$ & LLM orchestrator. \\
$\mathcal{I}_{(\cdot)}$ & Instruction prompt (select / miss / conf / update). \\
$\Phi(\cdot)$ & Backtracking decision function. \\
$\mathbb{I}[\cdot]$ & Indicator function. \\
$Imp(\cdot)$ & Importance function. \\
\bottomrule
\end{tabular}
\end{table}

\subsection{Framework Overview} \label{sec:architecture}
\begin{figure*}
    \centering
    \includegraphics[width=1\textwidth]{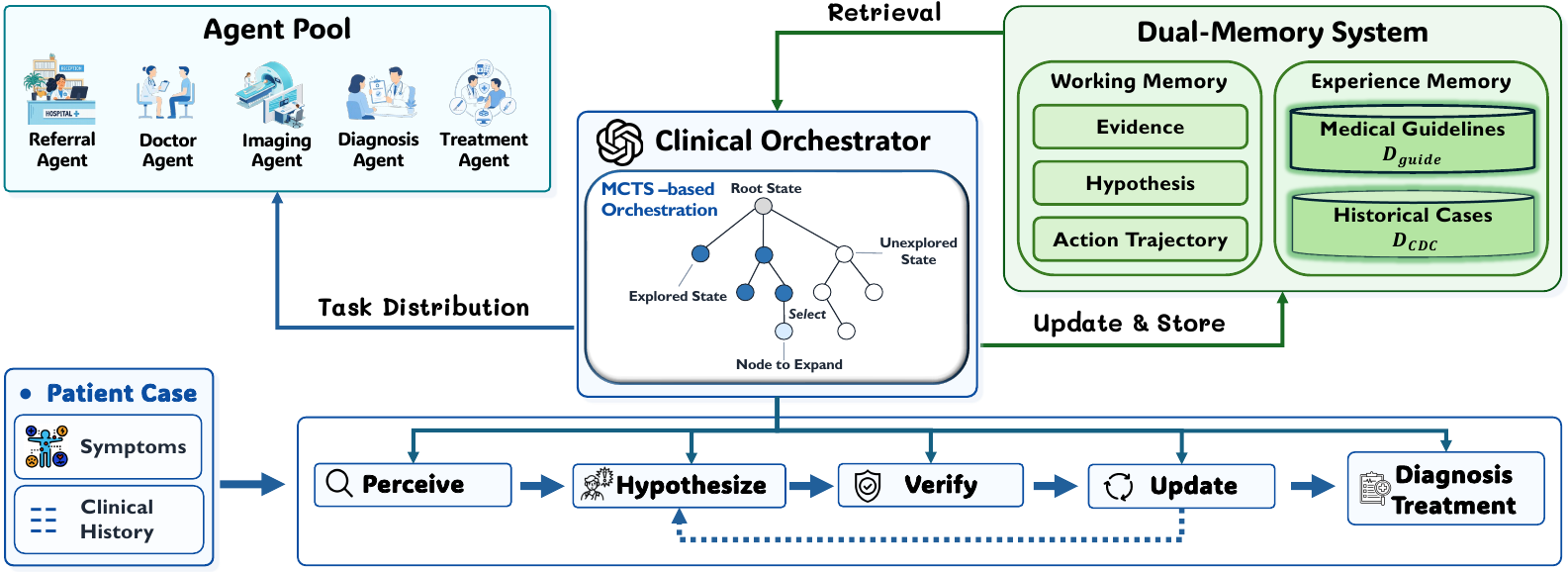}
    \caption{Overview of the ClinicalAgents framework. A Clinical Orchestrator coordinates a pool of specialized agents and a dual-memory system (working memory and experience memory) to execute an iterative Perceive--Hypothesize--Verify--Update loop with evidence-triggered backtracking.}
    \label{fig:pipeline}
\end{figure*}

The architecture of ClinicalAgents is built around a diverse array of specialized doctor agents that encompass the entire clinical workflow, from the initial patient visit to the final diagnosis.
As illustrated in Figure \ref{fig:pipeline}, the framework comprises three core components:
\begin{itemize}
    \item \textbf{Agent Pool}: This component contains specialized agents representing specific medical roles and tasks.
    \item \textbf{Clinical Orchestrator}: Serving as the executive controller, it manages task distribution, information flow, and the execution of backtracking procedures. Details are discussed in Sec.~\ref{sec:orchestrator}.
    \item \textbf{Dual-Memory System}: This system facilitates state synchronization across agents, consisting of a mutable \textit{Working Memory} $\mathcal{M}_\text{work}$ and a static \textit{Experience Memory} $\mathcal{M}_{\mathrm{exp}}$. Sec.~\ref{sec:memory} elaborates on its design and functionality.
\end{itemize}
The workflow follows a rigorous sequence: \textit{Perceive $\rightarrow$ Hypothesize $\rightarrow$ Verify $\rightarrow$ Update}. 
Within this cycle, the Clinical Orchestrator dynamically activates agents to generate candidate hypotheses and validates them against clinical guidelines. 
Crucially, this process is reinforced by an active \textbf{Backtracking Mechanism}: if the verification step reveals insufficient evidence, the system reverts to the perception phase to retrieve, inspect, or request the missing information from the case record, marking it as unresolved if it remains unavailable.
In the subsequent sections, we provide details on the design and implementation of these core modules.

\subsection{Clinical Orchestrator} \label{sec:orchestrator}
The Clinical Orchestrator functions as the central control unit of the ClinicalAgents framework, responsible for orchestrating the interactions and task execution across specialized agents. 
Overall, we formulate the agent orchestration as a memory-based Monte Carlo Tree Search (MCTS) procedure~\cite{silver2016mastering,browne2012survey}, with a hypothesis-driven expansion strategy and a backtracking mechanism to acquire missing evidence and produce a final diagnosis. 
In each expansion step of the MCTS, we carry out agent selection within the Markov Decision Process (MDP)~\cite{puterman1990markov}, where the orchestrator dynamically selects and activates the most suitable agents according to the patient's current status and evolving diagnostic requirements.
Specifically, we give our problem definition as follows:
	\begin{definition}[The Memory-based MCTS Problem]
		In each step $t$, given a Markov Decision Process (MDP) defined as the tuple $\langle \mathcal{M}_\text{work}^t, \mathcal{A}, \mathcal{R}, \mathcal{T}, \eta \rangle$, where:
		\begin{itemize}
			\item $ \mathcal{M}^t_\text{work}$ is the working memory at step $t$, which includes the evidence set $\mathcal{E}_t$, hypothesis set $\mathcal{H}_t$, and action trajectory $\tau_t=(a_0, a_1, \cdots, a_t)$:
			\begin{equation}\label{eq:memory}
				\mathcal{M}^t_\text{work}=\langle \mathcal{E}_t, \mathcal{H}_t, \tau_t \rangle
			\end{equation}
			\item $\mathcal{A}$ is the action space,
			\item $\mathcal{R}$ represents the reward function,
			\item $\mathcal{T}$ is the terminal state (e.g., clinical closure reached with a confirmed diagnosis),
			\item $\eta$ is the maximum orchestration step.
		\end{itemize}
		
The objective of our MCTS is to approximate the optimal policy $\pi^*: \mathcal{M}_\text{work}^t \to \mathcal{A}$ starting from step $t=0$, maximizing the expected {information gain} with reward function $\mathcal{R}$ (Eq.~\eqref{eq:dense_reward}).
The MDP ends when it reaches the terminal state $\mathcal{T}$, or reaches the maximum orchestration step $\eta$.
\end{definition}

	In each step $t$, given the working memory $\mathcal{M}_\text{work}^t$, the LLM orchestrator $\mathcal{O}$ estimates the probability of taking a specific action $a$ to advance the diagnosis with reward function $\mathcal{R}$.
	In the MDP procedure, the action space $\mathcal{A}$ includes:
	\begin{enumerate}[label=(\roman*), leftmargin=0pt, itemindent=*, align=left]
		\item $\mathcal{A}_{\text{agent}}$: It calls specific agents (e.g., image diagnosis agent, treatment agent, ...) to carry out certain tasks as needed.
		\item $a_{\text{rag}}$: It calls the knowledge retrieval action from the experience memory $\mathcal{M}_\text{exp}$, to provide new evidence or guidance.
		\item $a_{\text{back}}$: It calls the orchestrator $\mathcal{O}$ to carry out backtracking and decide the backtracking point in trajectory $\tau$.
		\item $a_{\text{term}}$: It terminates the orchestration process when clinical closure is reached.
	\end{enumerate}
	\begin{equation}
		\mathcal{A} = \mathcal{A}_{\text{agent}} \cup \{a_{\text{rag}}, a_{\text{back}}, a_{\text{term}}\}.
	\end{equation}
		With backtracking enabled, our orchestration is modeled as a searchable decision process, rather than hard-coded rules, allowing the system to revert to the perception phase when current evidence is insufficient to verify a hypothesis.

		Next, we introduce our reward function $\mathcal{R}$, which is based on the information gain after introducing each evidence item and making each hypothesis. 
		Let $\mathcal{E}_t^m$ be the set of missing critical evidence at step $t$, and let
		$c_t\in[0,1]$ be the confidence of the top hypothesis $h_t \in \mathcal{H}_t$ generated by the LLM.
		Define:
		\begin{equation}
			\label{eq:missing_and_confidence}
			\begin{split}
				\mathcal{E}_t^m = \text{LLM}(\mathcal{E}_t, \mathcal{I}_{\text{miss}}),\quad c_t = \text{LLM}(h_t, \mathcal{I}_{\text{conf}}), \\
				\Delta \mathcal{E}_t^m = \left|\mathcal{E}_{t-1}^m\right| - \left|\mathcal{E}_{t}^m\right|,\quad \Delta c_t = c_{t}-c_{t-1}.
			\end{split}
		\end{equation} 
		where $\mathcal{E}_t^m$ denotes the set of missing critical evidence and $c_t$ denotes the confidence score at step $t$, $\mathcal{E}_t$ and $\mathcal{H}_t$ denote the evidence set and hypothesis set at step $t$, $\mathcal{I}_{\text{miss}}$ and $\mathcal{I}_{\text{conf}}$ denote the missing evidence detection and confidence-scoring instruction. 
		To guide the search toward a verified conclusion, our reward function $\mathcal{R}$ mirrors the clinical goal of narrowing differential diagnoses, and provides positive feedback only when an action effectively reduces the evidence gap ($\Delta \mathcal{E}_t^m > 0$) or solidifies diagnostic confidence ($\Delta c_t > 0$). For any non-terminal action ($a_t \neq a_{\text{term}}$), the reward is:
		\begin{equation}
			\label{eq:dense_reward}
			\mathcal{R}_t
			= \alpha\cdot \max(0,\Delta \mathcal{E}_t^m)
			+ \beta \cdot \max(0,\Delta c_t)
			- \gamma \cdot \mathbb{I}\!\left[\Delta \mathcal{E}_t^m \le 0\ \wedge\ \Delta c_t \le 0\right],
		\end{equation}
		where $\alpha,\beta,\gamma \ge 0$ are weighting coefficients for missing-evidence reduction, confidence improvement, and uninformative-action penalty, respectively; $\Delta \mathcal{E}_t^m$ and $\Delta c_t$ denote the change in missing evidence and hypothesis confidence at step $t$ (Eq.~\eqref{eq:missing_and_confidence}). 
		The search concludes when the terminal action $a_\text{term}$ is selected, which denotes that the {clinical closure} is reached via a confirmed diagnosis and treatment plan; or when the maximum orchestration length $\eta$ is reached without resolving diagnostic uncertainty.
		
		With the action space $\mathcal{A}$ and reward function $\mathcal{R}$ defined, we describe the decision-making process of the orchestrator $\mathcal{O}$ as follows. 
		Starting from an initial $\mathcal{M}_\text{work}^t$ (the current working memory snapshot), the orchestrator repeatedly:
		\begin{enumerate}[label=(\roman*), leftmargin=0pt, itemindent=*, align=left]
			\item Selects an action $a_t\in\mathcal{A}$;
			\item Samples the next $\mathcal{M}_\text{work}^{t+1}$ with the orchestrator $\mathcal{O}$ considering the current $\mathcal{M}_\text{work}^t$ and action $a_t$, by executing the chosen agent/tool call and updating the working memory and workflow stage;
			\item Receives rewards with reward function $\mathcal{R}$, shaped to encourage evidence completion and increased top-hypothesis confidence.
			\item Repeats (i)-(iii) until the orchestration terminates.
		\end{enumerate}
		
	Next, we introduce our MCTS-based Expansion and Backtracking strategy. 
	Given the current $\mathcal{M}_\text{work}^t$, the orchestrator model $\mathcal{O}$ outputs an action selection score distribution $\mathcal{O}(a\mid \mathcal{M}_\text{work}^t,\mathcal{I}_{\text{select}})$ over all actions $a\in\mathcal{A}$.
	We first construct a candidate set by taking the top-$K$ actions:
	\begin{equation}
		\mathcal{A}_K=\operatorname{TopK}_{a\in\mathcal{A}}\ \mathcal{O}(a\mid \mathcal{M}_\text{work}^t,\mathcal{I}_{\text{select}}).
		\label{eq:topk}
	\end{equation}
	For each candidate action $a\in\mathcal{A}_K$, we run $N$ independent rollouts and estimate its action value $Q(\mathcal{M}_\text{work}^t,a)$ at step $t$ by the average rollout return:
	\begin{equation}
		\label{eq:q_value}
		Q(\mathcal{M}_\text{work}^t,a) = \frac{1}{N} \sum_{i=1}^{N} \sum_{k=t}^{L_i} \gamma_d^{\,k-t}\, \mathcal{R}_k^{(i)}, \quad a\in\mathcal{A}_K,
	\end{equation}
	where $L_i$ is the terminal step of the $i$-th rollout and $\mathcal{R}_k^{(i)}$ is the reward obtained at step $k$ in that rollout, and $\gamma_d \in (0,1]$ is a discount factor that bounds the cumulative return and prevents bias toward longer rollouts.
	For each candidate action $a$, each rollout is initialized by taking $a$ at step $t$.
	% where $\tau_{i}$ denotes the trajectory of the $i$-th MCTS simulation, $\gamma$ and $\mathcal{R}_n$ denotes the reward discount factor and reward at step $n$.
	During tree traversal, node selection is governed by the LLM-prior-guided Predictor and Upper Confidence bound for Trees (PUCT), which selects an action $a^*$ at step $t$: 
	\begin{equation}
		a^* = \arg\max_{a \in \mathcal{A}_K} \left( Q(\mathcal{M}^t_\text{work}, a) + \lambda \cdot \mathcal{O}(a \mid \mathcal{M}^t_\text{work},\mathcal{I}_{\text{select}}) \right),
	\end{equation}
	where $\lambda>0$ is a balance factor.

Inspired by MCTS backpropagation, we design a novel backtracking mechanism in our orchestration that serves as a dynamic corrective measure to address diagnostic uncertainty arising from insufficient evidence.
		
First, after each hypothesis generation, the generated hypotheses and their respective evidence are updated in the working memory $\mathcal{M}_\text{work}^t$, and then the orchestrator model verifies whether the current evidence is adequate to support the hypothesis based on $\mathcal{M}_\text{work}^t$. 
Here, with the missing key evidence set at step $t$ as $\mathcal{E}_t^m$ in Eq. \eqref{eq:missing_and_confidence}, if critical evidence is absent (i.e., $\mathcal{E}_t^m \neq \emptyset$), the system identifies these gaps and triggers the backtracking process.
This behavior is mechanistically driven by the decision function $\Phi(\mathcal{M}_\text{work}^t)$: 
		\begin{equation}
			\Phi(\mathcal{M}^t_\text{work}) = \mathbb{I}\left[ |\mathcal{E}_t^m| > 0 \right] = 
			\begin{cases} 
				1 , & \text{if } \mathcal{E}_t^m \neq \emptyset \\
				0 , & \text{otherwise}
			\end{cases}
		\end{equation}

Next, if the evidence is insufficient (e.g., a missing CT result or a crucial symptom), our backtracking mechanism is triggered, and the orchestrator $\mathcal{O}$ decides where to resume the workflow. 
The key idea is that different missing evidence types imply different task-stage rollbacks: for example, missing imaging results should route the process back to an imaging interpretation stage, while missing history/symptoms should route back to the information-gathering stage. 
		Formally, conditioned on the missing-evidence $\mathcal{E}_{t}^m$ at step $t$ and the updated working memory $\mathcal{M}_\text{work}^{t+1}$, the orchestrator $\mathcal{O}$ selects the backtracking target stage and action $a_{\text{call}}$ (action in $\mathcal{A}$):
		\begin{equation}
			\begin{split}
				\mathcal{M}_\text{work}^{t+1} \leftarrow \mathcal{O}(\mathcal{M}^t_\text{work}, \mathcal{E}_{t}^m, \mathcal{I}_{\text{update}}), \\
				a_{\text{call}} \sim {\mathcal{O}}(\cdot \mid \mathcal{M}_\text{work}^{t+1}, \mathcal{I}_{\text{select}}).
			\end{split}
		\end{equation}
		
% The selected agent $a_{\text{call}}$ is then activated to continue hypothesis refinement under the newly completed evidence. 
% This design makes backtracking a {structured, evidence-driven backtracking} rather than a fixed rule, enabling the orchestrator to adaptively revisit the most relevant stage to resolve diagnostic uncertainty.

\subsection{Dual-Memory System} \label{sec:memory}
To support consistent multi-turn reasoning, ClinicalAgents adopts a dual-memory architecture $\langle\mathcal{M}_\text{work}, \mathcal{M}_\text{exp}\rangle$. Here, the working memory $\mathcal{M}_\text{work}$ maintains the evolving sets of evidence and hypotheses throughout the reasoning process, while the experience memory $\mathcal{M}_\text{exp}$ provides access to external medical guidelines and historical diagnostic cases. This design enables ClinicalAgents to integrate real-time patient context with established clinical knowledge, ensuring logical coherence across multiple reasoning steps. We detail the structure and function of each memory below.

\subsubsection{Working Memory} \label{sec:working-memory}
The working memory $\mathcal{M}_\text{work}$ acts as the central state board for the entire ClinicalAgents system, synchronizing all agents’ understanding of the current patient context and diagnostic progress. By aggregating and sharing the latest evidence and hypotheses, it ensures that all agent actions remain consistent and informed throughout multi-turn reasoning.
Formally, the working memory at step $t-1$ is defined as: $\mathcal{M}_\text{work}^{t-1}=\langle \mathcal{E}_{t-1},\mathcal{H}_{t-1},\tau_{t-1}\rangle,$
where $\mathcal{E}_{t-1}$ is the set of accumulated evidence (such as symptoms and test results), $\mathcal{H}_{t-1}$ is the evolving set of diagnostic hypotheses, and $\tau_{t-1}$ records the sequence of actions taken up to step ${t-1}$.
At each step $t$, the orchestrator updates the working memory by incorporating new evidence $\mathcal{E}^\text{new}_{t}$ and new hypotheses $\mathcal{H}^\text{new}_{t}$ collected from the agent interactions:
\[
	\mathcal{E}_{t}=\mathcal{E}_{t-1}\cup \mathcal{E}^\text{new}_{t}, \quad \mathcal{H}_{t}=\mathrm{UpdateHyp}(\mathcal{H}_{t-1}, \mathcal{H}_t^\text{new}).
\]
This design enables real-time, multi-turn synchronization of all agents, providing a comprehensive and continuously updated context for coherent clinical reasoning and decision-making.

\subsubsection{Experience Memory} \label{sec:experience-memory}
The {experience memory} $\mathcal{M}_\mathrm{exp}$ equips ClinicalAgents with external, context-aware knowledge that goes beyond the immediate patient state. 
Formally, the experience memory is defined as a static external memory store:
\[
\mathcal{M}_\mathrm{exp}
=
\left\langle 
\mathcal{D}_\mathrm{guide},\, \mathcal{D}_\mathrm{cdc}
\right\rangle,
\]
where $\mathcal{D}_\mathrm{guide}$ denotes the structured guideline database and 
$\mathcal{D}_\mathrm{cdc}$ denotes the historical case database.

At each reasoning step $t$, given the accumulated evidence $\mathcal{E}_t$ and the evolving diagnostic hypotheses $\mathcal{H}_t$, experience memory supports the system in two main ways: (1) by retrieving relevant clinical knowledge $\mathcal{K}_t$ from medical guidelines $\mathcal{D}_\mathrm{guide}$, and (2) by identifying potentially missing but clinically important evidence $\mathcal{E}_t^{p}$ for the current diagnostic context using the historical case database $\mathcal{D}_\mathrm{cdc}$.
The experience memory produces step-specific retrieval outputs:
\[
(\mathcal{K}_t,\, \mathcal{E}_t^{p})
=
\mathrm{RetrieveExp}
\big(
\mathcal{M}_\mathrm{exp},\, \mathcal{E}_t,\, \mathcal{H}_t
\big).
\]

First, for the retrieval of guideline-based knowledge, given the structured guideline database $\mathcal{D}_\mathrm{guide}$ (including authoritative recommendations, evidence-based protocols, and diagnostic pathways), along with the current evidence $\mathcal{E}_t$ and diagnostic hypotheses $\mathcal{H}_t$, the system retrieves relevant knowledge as
\[
\mathcal{K}_t = \mathrm{Retriever}\big(\mathcal{E}_t \oplus \mathcal{H}_t,\, \mathcal{D}_\mathrm{guide}, n_\mathrm{guide}\big),
\]
where $\mathrm{Retriever}$ refers to our previously proposed retrieval model AGRAG~\cite{wang2025agrag}, $n_{\mathrm{guide}}$ denotes the number of retrieved guideline chunks, and $\oplus$ denotes the concatenation of evidence and hypotheses to form the input query.
AGRAG combines chunk-level retrieval with knowledge-graph-based entity importance, enabling the system to capture not only contextual relevance but also the clinical significance of retrieved entities, ensuring comprehensive and meaningful guidance extraction.

Second, for identifying missing potential evidence, the system utilizes $\mathcal{D}_\mathrm{cdc}=\{d_i\}_{i=1}^{|\mathcal{D}_\mathrm{cdc}|}$, a repository of historical clinical cases. 
Each case $d_i = \langle \mathcal{E}^*_{d_i},\, \mathcal{H}^*_{d_i},\, \mathcal{P}^*_{d_i} \rangle \in \mathcal{D}_\mathrm{cdc}$ is represented as a Causal Diagnostic Chain (CDC), where $\mathcal{E}^*_{d_i}$ denotes the diagnostic evidence, $\mathcal{H}^*_{d_i}$ is the diagnosis, and $\mathcal{P}^*_{d_i}$ specifies the treatment plan.
Given the current evidence $\mathcal{E}_t$ and diagnostic hypotheses $\mathcal{H}_t$, the system retrieves the top-$n_\mathrm{cdc}$ most similar historical cases,
\[
\mathcal{\tilde{D}}_\mathrm{cdc}= \mathrm{Retriever}\big(\mathcal{E}_t \oplus \mathcal{H}_t,\, \mathcal{D}_\mathrm{cdc}, n_\mathrm{cdc}\big),
\]
where $\oplus$ denotes concatenation for query formation and $n_{\mathrm{cdc}}$ denotes the number of retrieved clinical cases. 

Then, we identify key evidence items likely to be missing from the current diagnostic context $\mathcal{E}_t$ based on the retrieved clinical case set $\mathcal{\tilde{D}}_\mathrm{cdc}$.

Next, with $\tilde{\mathcal{D}}_\mathrm{cdc}$, we take the union of their key evidence fields to form $\tilde{\mathcal{E}}_\mathrm{cdc} = \bigcup_{d_i \in \tilde{\mathcal{D}}_\mathrm{cdc}, e \in \mathcal{E}^*_{d_i}}{e}$, 
where $e \in \mathcal{E}^*_{d_i}$ denotes a key evidence item $e$ in case $d_i$. 
Then, for each $e \in \tilde{\mathcal{E}}_\mathrm{cdc}$, the system computes an importance score as follows:
\[
Imp(e|\mathcal{E}_t,\mathcal{\tilde{D}}_\mathrm{cdc},\mathcal{H}_t) = \frac{1}{n_{\mathrm{cdc}}} \sum_{i=1}^{n_{\mathrm{cdc}}} \mathrm{sim}(\mathcal{E}_t \oplus \mathcal{H}_t,\, d_i) 
\cdot \mathbb{I}[e \in \mathcal{E}^*_{d_i}] \cdot \mathbb{I}[e \notin \mathcal{E}_t],
\]
where $\mathrm{sim}(\cdot)$ is {a cosine similarity metric}, and $\mathbb{I}[\cdot]$ is the indicator function, returning $1$ if the condition is satisfied and $0$ otherwise. 

% Next, with $\tilde{\mathcal{D}}_\mathrm{cdc}$, we take the union of their key evidence fields and exclude items already observed in $\mathcal{E}_t$ to form the candidate evidence set
% \[
% \tilde{\mathcal{E}}_\mathrm{cdc} = \Big(\bigcup_{d_i \in \tilde{\mathcal{D}}_\mathrm{cdc}} \mathcal{E}^*_{d_i}\Big) \setminus \mathcal{E}_t,
% \]
% where $\mathcal{E}^*_{d_i}$ denotes the set of key evidence items in case $d_i$.
% Then, for each $e \in \tilde{\mathcal{E}}_\mathrm{cdc}$, the system computes an importance score as follows:
% \[
% Imp(e\mid\mathcal{E}_t,\tilde{\mathcal{D}}_\mathrm{cdc},\mathcal{H}_t) = \frac{1}{n_{cdc}} \sum_{d_i \in \tilde{\mathcal{D}}_\mathrm{cdc}} \mathrm{sim}(\mathcal{E}_t \oplus \mathcal{H}_t,\, d_i) 
% \cdot \mathbb{I}[e \in \mathcal{E}^*_{d_i}],
% \]
% where $\mathrm{sim}(\cdot)$ is a cosine similarity metric, and $\mathbb{I}[\cdot]$ is the indicator function, returning $1$ if the condition is satisfied and $0$ otherwise.

The set of missing but clinically significant evidence is then determined by
\[
 \mathcal{E}_t^{p} = \left\{ e \in \tilde{\mathcal{E}}_\mathrm{cdc} \mid Imp(e|\mathcal{E}_t,\mathcal{\tilde{D}}_\mathrm{cdc},\mathcal{H}_t) > \delta \right\},
\]
where $\delta$ is a predefined significance threshold. This approach enables the system to proactively identify and highlight crucial, yet previously overlooked, evidence items, thereby supporting more robust and context-aware diagnostic reasoning.

% \clearpage
 \renewcommand{\arraystretch}{1.2}

 \renewcommand{\arraystretch}{1.0}
\section{Experiments} \label{sec:experiment}
In this section, we systematically evaluate ClinicalAgents' performance on multi-stage clinical workflow reasoning tasks and compare it with strong baselines, including general LLM/VLM, single-agent systems, and representative multi-agent systems. 
We also conduct ablation of key components, sensitivity experiments on different backbone models, and an efficiency analysis against test-time compute baselines.
Specifically, we explore the following research questions (\textbf{RQs}):
\begin{itemize}
    \item \textbf{RQ1:} How does ClinicalAgents perform compared to strong baselines on the MedChain benchmark~\cite{liu2024medchain}? (Sec. \ref{sec:results})
    \item \textbf{RQ2:} What is the contribution of each component of ClinicalAgents to the overall performance? (Sec. \ref{sec:ablation})
    \item \textbf{RQ3:} How sensitive is ClinicalAgents to different backbone LLMs? (Sec. \ref{sec:parameter})
    \item \textbf{RQ4:} Do the gains of ClinicalAgents arise from multi-agent orchestration rather than from larger test-time compute budgets? (Sec.~\ref{sec:efficiency})
    % \item \textbf{RQ4:} Can ClinicalAgents generate coherent and clinically relevant reasoning paths in complex medical cases? (Sec. \ref{sec:case_study})
    % \item \textbf{RQ5:} How does ClinicalAgents perform on downstream medical QA tasks? (Sec. \ref{sec:qa})
\end{itemize}

\subsection{Experimental Settings} \label{sec:settings}
In this subsection, we describe the experimental settings, including the datasets and their corresponding tasks, the evaluation metrics, the baseline methods for comparison, and the overall implementation details of our experiments.

\subsubsection{Datasets and Evaluation Metrics} \label{sec:data_metrics}
We primarily evaluate on MedChain~\cite{liu2024medchain}, and further test generalization on MedDG~\cite{liu2022meddg} and four medical QA datasets.
Dataset statistics are summarized in Table~\ref{tab:dataset_stats}.
Detailed descriptions of the MedDG and four QA datasets are provided in Appendix~\ref{exp:qa}.

\paragraph{MedChain (Benchmark).}
MedChain~\cite{liu2024medchain} is organized as a sequential, interdependent clinical workflow:
\[
\text{Task 1} \rightarrow \text{Task 2} \rightarrow \text{Task 3} \rightarrow \text{Task 4} \rightarrow \text{Task 5}.
\]
Outputs from earlier stages are explicitly fed into subsequent ones, enforcing \textbf{cross-stage correlation} and \textbf{sequential consistency} across the pipeline.
Concretely, the five stage-level tasks and their evaluation metrics are:
\begin{itemize}
  \item \textbf{Task 1: Specialty Referral} determines case urgency and routes the patient to an appropriate clinical department.
  We use \textbf{Accuracy} for first-level department and \textbf{Intersection over Union (IoU)} for second-level department assignment.

  \item \textbf{Task 2: Test Ordering} specifies examinations/tests to be ordered.
  We use \textbf{IoU} to measure set overlap between predicted and reference medical items (coverage and precision).

  \item \textbf{Task 3: Examination} summarizes the results of requested examinations, including medical image interpretation and report drafting.
  We use \textbf{DocLens}~\cite{xie2024doclens} to assess the quality of generated image interpretation compared to references.

  \item \textbf{Task 4: Diagnosis} integrates evidence from Tasks~1--3 to produce a diagnostic conclusion.
  We use \textbf{LLM-as-a-Judge} to evaluate diagnostic correctness and reasoning coherence.

  \item \textbf{Task 5: Treatment} formulates a patient-specific treatment plan under clinical constraints.
  We use \textbf{IoU} to measure overlap between predicted and reference treatment items.
\end{itemize}

 \begin{table}[t]
	\centering
	\caption{Dataset statistics in our experiments. T=Text, I=Image; “5-step pipeline” denotes a five-stage workflow.}
	\label{tab:dataset_stats}
	\small
	\setlength{\tabcolsep}{5pt} % Adjust column separation
	\begin{tabular}{lccc}
		\toprule
		\textbf{Dataset} & \textbf{\# Samples} & \textbf{Modality} & \textbf{Task} \\
		\midrule
		MedChain~\cite{liu2024medchain} & 12{,}163 & T+I & 5-stage pipeline \\
		% MedChain (test) & 2{,}023 & T+I & 5-stage pipeline \\
		% \midrule
		MedDG~\cite{liu2022meddg} & 17{,}864 & T & dialogue generation \\
		MedQA~\cite{jin2021disease} & 1{,}273 & T & multiple QA \\
		PubMedQA~\cite{jin2019pubmedqa} & 500 & T & y/n/maybe QA \\
		MedBullets~\cite{chen2025benchmarking} & 308 & T & multiple QA \\
		PathVQA~\cite{he2020pathvqa} & 3{,}362 & I & binary QA \\
		\bottomrule
	\end{tabular}
	\vspace{-5px}
\end{table}

Since Task~3 involves medical image interpretation, we focus on the \textbf{image-containing subset} (3{,}034 cases) to assess multimodal clinical reasoning.
From this subset, we randomly sample two-thirds of the cases (2{,}023) for the main experiments using a fixed random seed (seed=42).
For text-only cases, we use them as historical cases in the experience memory, and the test cases are strictly held out from the experience memory to prevent data leakage.

\subsubsection{Baselines} \label{sec:baselines}
Our evaluation is designed to be comprehensive, covering increasingly structured paradigms from single LLMs to single-agent prompting/tool-use and finally to multi-agent systems. 

\begin{itemize}
    \item \textbf{Five large language models:} We include both closed-source and open-source models to balance strong proprietary systems with reproducible open models, and to cover text-only and multimodal capabilities.
    The closed-source model is GPT-5.2~\cite{achiam2023gpt}, while the open-source models include Qwen3-VL-235B~\cite{bai2025qwen3vltechnicalreport}, DeepSeek-R1~\cite{guo2025deepseek}, Llama-4-Maverick~\cite{touvron2023llama}, and Intern-S1~\cite{chen2024internvl,bai2025intern}.
    \item \textbf{Three fine-tuned medical LLMs:} We additionally evaluate domain-specialized models that are explicitly trained or fine-tuned for medical reasoning and QA, serving as strong in-domain baselines: MedGemma-4B~\cite{sellergren2025medgemma}, HuaTuoGPT-o1-7B~\cite{zhang2023huatuogpt,chen2024huatuogpt}, and FineMedLM-o1-8B~\cite{yu2025finemedlm}.
    \item \textbf{Single-agent:} We compare ClinicalAgents with widely adopted single-agent paradigms that progressively increase structure and external knowledge usage, including: Few-shot + CoT~\cite{wei2022chain}, RAG and ReAct~\cite{yao2022react}.
    \item \textbf{Multi-agent systems:} To benchmark orchestration and coordination benefits, we include representative multi-agent frameworks that have been used across reasoning and domain tasks: ReConcile~\cite{chen2024reconcile}, AutoGen~\cite{wu2024autogen}, MedAgents~\cite{tang2024medagents}, MDAgents~\cite{kim2024mdagents}, and ColaCare~\cite{wang2025colacare}, together with MedChain-Agents~\cite{liu2024medchain}, a strong task-specific baseline on the MedChain benchmark.
\end{itemize}

We select baselines using three criteria: (i) \textbf{representativeness} (widely used or commonly adopted in prior work), (ii) \textbf{strength} (competitive or state-of-the-art performance in general or medical domains), and (iii) \textbf{coverage} (spanning closed/open models, text-only and multimodal settings).

\subsubsection{Implementation Details} \label{sec:implementation}
We implement all methods in a unified pipeline to ensure consistent preprocessing, prompt formatting, and evaluation.
\begin{itemize}
\item \textbf{Backbone models.} Unless otherwise specified, we use GPT-5.2 as the backbone for all single-agent and multi-agent baselines, as well as ClinicalAgents, to ensure a fair comparison.
\item \textbf{LLM-as-a-Judge.} We use Gemini-2.5-Pro (temperature $=0$) to score diagnostic correctness and reasoning coherence on an integer scale of $1$ to $5$, with a rubric grounded in established clinical diagnostic principles that tolerates synonymous expressions and reasonable granularity differences. The full prompt and rubric are provided in Appendix~\ref{prompts:agents}.
%   \item \textbf{Prompts.} To facilitate future research and ensure experimental reproducibility, the specific prompt templates used for each task are comprehensively detailed in Appendix A.
\item \textbf{Parameters.} We set the maximum orchestration length to $\eta=4$ and enable early stopping when the orchestrator reaches a confident diagnosis or selects the terminal action. The full configuration file will be released with the code repository.
\end{itemize}

\begin{table*}[htbp]
	\caption{Overall results on the MedChain benchmark. We compare ClinicalAgents with single-LLM, single-agent, and multi-agent baselines across five tasks representing workflow stages, and report score for each task and the average score. The symbol `—' indicates that the model is a text-only LLM and not applicable to a visual examination task, and average score is computed over applicable non-visual metrics only. The LLM-judge score was normalized to 0–1 by dividing by 5 before averaging.}
	\centering
	% \caption{Overall results on 5 tasks of MedChain benchmark.}
	\label{tab:medchain_main}
	\small
	\setlength{\tabcolsep}{4pt}
	\renewcommand{\arraystretch}{1.1}
	\begin{tabular}{llccccccc}
		\toprule
		\multirow{3}{*}{\textbf{Category}} &
		\multirow{3}{*}{\textbf{Methods}} &
		\multicolumn{6}{c}{\textbf{MedChain}} &
		\multirow{3}{*}{\textbf{Average}} \\
		\cmidrule(lr){3-8}
		& & \multicolumn{2}{c}{\textbf{Specialty Referral}} & \textbf{Test Ordering} & \textbf{Examination} & \textbf{Diagnosis} & \textbf{Treatment} & \\
		\cmidrule(lr){3-4}
		& & \textbf{Acc. (Lv1)} & \textbf{IoU (Lv2)} & \textbf{IoU} & \textbf{DocLens} & \textbf{LLM-judge} & \textbf{IoU} & \\
		\midrule
		
		\multirow{9}{*}{\textbf{Single LLM}}
		& GPT-5.2                     & 0.5467 & 0.3045 & 0.3870 & 0.4276 & 0.5754 & 0.4716 & 0.4521 \\
		& DeepSeek-R1                  & 0.5744 & 0.3124 & 0.3475 & \textemdash & 0.4968 & 0.4296 & 0.4321 \\
		& Qwen3-VL-235B                & 0.5719 & \textbf{0.3604} & 0.2867 & 0.3915 & 0.5798 & 0.5017 & 0.4487 \\
		& Llama-4-Maverick-17B         & 0.5620 & 0.3386 & 0.3075 & 0.3579 & 0.5675 & 0.4048 & 0.4231 \\
		& Intern-S1                    & 0.5783 & 0.3406 & 0.3317 & 0.4192 & 0.4810 & 0.3964 & 0.4245 \\
		& MedGemma-4B                  & 0.4014 & 0.2477 & 0.2630 & 0.4063 & 0.5467 & 0.4380 & 0.3838 \\
		& HuaTuoGPT-o1-7B              & 0.3609 & 0.2037 & 0.1537 & \textemdash & 0.5442 & 0.4647 & 0.3454 \\
		& FineMedLM-o1-8B              & 0.3688 & 0.1948 & 0.2140 & \textemdash & 0.4716 & 0.4157 & 0.3330 \\
		\midrule
		
		\multirow{2}{*}{\textbf{Single Agent}}
		& Few-shot + CoT               & 0.5680 & 0.3047 & 0.3886 & 0.4434 & 0.5762 & 0.4839 & 0.4608 \\
		& RAG                          & 0.5675 & 0.3041 & 0.3789 & 0.4499 & 0.5781 & 0.4876 & 0.4610 \\
		& ReAct~\cite{yao2022react}     & 0.5803 & 0.3212 & 0.3961 & 0.4543 & 0.5820 & 0.4835 & 0.4696 \\
		\midrule
		
		\multirow{7}{*}{\textbf{Multi-Agents}}
		& ReConcile~\cite{chen2024reconcile}                & 0.5882 & 0.3431 & 0.4026 & 0.4790 & 0.5774 & 0.4869 & 0.4795 \\
		& AutoGen~\cite{wu2024autogen}                      & 0.5615 & 0.3243 & 0.4156 & 0.4404 & 0.5764 & 0.5022 & 0.4701 \\
		& MedAgents~\cite{tang2024medagents}                & 0.5344 & 0.2921 & 0.3961 & 0.4676 & 0.5442 & 0.4938 & 0.4547 \\
		& MDAgents~\cite{kim2024mdagents}                   & 0.5442 & 0.2887 & 0.4072 & 0.4582 & 0.5759 & 0.4825 & 0.4594 \\
		& ColaCare~\cite{wang2025colacare}                  & 0.5798 & 0.3273 & 0.3994 & 0.4441 & 0.5623 & 0.4920 & 0.4675 \\
		& MedChain-Agents~\cite{liu2024medchain}            & 0.5937 & 0.3531 & 0.4382 & 0.4528 & 0.5863 & 0.5041 & 0.4880 \\
		% \rowcolor{gray!15}
		& \textbf{Ours} & \textbf{0.6243} & 0.3543 & \textbf{0.4820} & \textbf{0.4800} & \textbf{0.5976} & \textbf{0.5259} & \textbf{0.5107} \\
		
		\bottomrule
	\end{tabular}
\end{table*}

\subsection{Overall Results (RQ1)} \label{sec:results}
In this subsection, we present the overall performance of ClinicalAgents compared to strong baselines on the MedChain benchmark~\cite{liu2024medchain}, addressing \textbf{RQ1}. 
To assess generalization beyond the MedChain workflow setting, we additionally conduct experiments on multiple medical dialogue and QA datasets; these results are reported in Appendix~\ref{exp:qa}.

The main experimental results are reported in Table~\ref{tab:medchain_main}.
Overall, our method ClinicalAgents achieves the best performance among all compared approaches, reaching an average score of 0.5107. 
% This result consistently surpasses strong baselines across single LLM, single agent, and multi-agent settings. 
Notably, ClinicalAgents attains the best score on 5 out of 6 reported metrics, indicating robust improvements across different stages of the MedChain benchmark~\cite{liu2024medchain}. 
Specifically, we provide the following detailed observations and analysis:
\begin{itemize}
    \item \textbf{ClinicalAgents achieves the best performance with an average score of 0.5107.} 
    Compared with the backbone GPT-5.2 (0.4521), ClinicalAgents improves the average score by +0.0586, corresponding to a +13.0\% relative gain. 
    In addition, ClinicalAgents consistently outperforms strong agentic baselines: it exceeds the best multi-agent baseline MedChain-Agents (0.4880) by +0.0227 (+4.7\%), and the strongest single-agent baseline ReAct (0.4696) by +0.0411 (+8.8\%).
    This substantial improvement suggests that ClinicalAgents effectively enhances clinical reasoning through the orchestrator’s backtracking mechanism, reducing error accumulation and improving cross-stage consistency.  
    \item \textbf{ClinicalAgents delivers particularly strong gains on Task 2 (Test Ordering), achieving the best alignment with the ground truth.}
    ClinicalAgents attains a Test Ordering IoU of 0.4820, ranking first among all methods.
    It improves over the backbone GPT-5.2 (0.3870) by {+0.0950} ({+24.5\%}), and surpasses the strongest baseline MedChain-Agents (0.4382) by {+0.0438} ({+10.0\%}).
    We attribute this gain to the orchestrator’s verification-and-backtracking mechanism, which iteratively queries for missing evidence and refines the examination plan by adding critical tests while removing redundant or low-yield items.
    \item 
    \textbf{ClinicalAgents achieves performance gains even on the vision-based Examination task.}
	Although ClinicalAgents' visual processing primarily depends on the visual capability of its backbone LLM, it still achieves performance gains on the Examination task (Task 3), which requires visual understanding. 
	The improvement mainly stems from hypothesis-driven examination selection: ClinicalAgents leverages current hypotheses and experience memory to select targeted examinations, and subsequently injects these hypotheses into the imaging-agent prompt as clinically relevant targets to guide the model's attention toward findings associated with each candidate diagnosis.
\end{itemize}

\paragraph{Reliability of LLM-as-a-Judge.}
Because Task~4 relies on an LLM judge, we validated its reliability against human experts. 
We conducted a human study on a 20\% sample of the held-out set ($N=202$ out of $1{,}011$ cases not used in the main experiments), with two clinicians independently scoring each case using the same 1--5 rubric.
As shown in Table~\ref{tab:llm_judge_agreement}, the LLM judge achieves strong agreement with both clinicians (Cohen's $\kappa \approx 0.80$ and within-one-point Tolerance Accuracy $\geq 88\%$), comparable to the inter-clinician agreement, supporting its validity as an evaluation proxy.
 
\begin{table}[t]
\centering
\small
\setlength{\tabcolsep}{8pt}
\caption{Agreement between the LLM judge (Gemini~2.5~Pro) and two clinicians on the Task~4 diagnostic score. Cohen's $\kappa$ corrects for chance agreement; Tolerance Accuracy ($\Delta\!\leq\!1$) is the percentage of cases where two raters' scores differ by no more than one point on the 1--5 scale.}
\label{tab:llm_judge_agreement}
\begin{tabular}{lcc}
\toprule
\textbf{Agreement Pair} & \textbf{Cohen's $\kappa$} & \textbf{Tol. Acc. ($\Delta\!\leq\!1$)} \\
\midrule
Clinician A vs. LLM Judge & 0.77 & 88\% \\
Clinician B vs. LLM Judge & 0.82 & 92\% \\
Clinician A vs. Clinician B & 0.88 & 94\% \\
\bottomrule
\end{tabular}
\end{table}

\paragraph{Reliability of the backtracking mechanism.}
To probe whether the orchestrator can detect and recover from initially incorrect but plausible hypotheses, we compute three reliability statistics over our experiments (Table~\ref{tab:backtracking_stats}). 
When the initial hypothesis is wrong, the orchestrator successfully triggers backtracking in 83\% of such cases and ultimately recovers the correct diagnosis in 64\%. 
Across all cases, at least one unsupported hypothesis is explicitly pruned 94\% of the time, showing active hypothesis exclusion rather than uniform expansion. 
Detailed analysis of case studies of the backtracking mechanism is provided in Appendix~\ref{sec:case_study}.
% The remaining failures concentrate on atypical symptom presentations or cases with missing crucial clinical evidence; representative success and failure case studies, along with a detailed failure-mode analysis, are provided in Appendix~\ref{sec:case_study}.
 
\begin{table}[t]
\centering
\small
\setlength{\tabcolsep}{8pt}
\caption{Reliability statistics of the orchestrator's backtracking mechanism. The trigger and recovery rates are computed over cases whose initial hypothesis was judged incorrect.}
\label{tab:backtracking_stats}
\begin{tabular}{lr}
\toprule
\textbf{Metric} & \textbf{Value} \\
\midrule
Backtracking trigger rate (initially wrong cases) & 83\% \\
Recovery success rate (initially wrong cases)     & 64\% \\
Cases with at least one excluded hypothesis       & 94\% \\
\bottomrule
\end{tabular}
\end{table}

\subsection{Ablation Study (RQ2)} \label{sec:ablation}
In this subsection, we assess the contribution of each key component to address \textbf{RQ2}.
We isolate the marginal effect of the \textit{Dual-Memory} (working memory + experience memory) and the \textit{Clinical Orchestrator} (MCTS-based action selection with backtracking) via four configurations:
\begin{itemize}
    \item \textbf{Backbone}: a fixed five-stage pipeline with standard prompt chaining. It does not use working memory, experience memory, or MCTS orchestration. It remains functional because each stage's output is passed directly to the next stage.
    \item \textbf{+ Dual-Memory}: Backbone augmented with working memory and experience memory; the workflow remains fixed (no MCTS, no backtracking).
    \item \textbf{+ Orchestrator}: Backbone augmented with working memory and the MCTS-based orchestrator (with backtracking), but \emph{without} experience memory.
    \item \textbf{ALL}: the full ClinicalAgents with working memory, experience memory, and the MCTS-based orchestrator.
\end{itemize}
As shown in Table~\ref{tab:ablation_medchain}, both Dual-Memory and the orchestrator independently yield substantial gains over the backbone model. 
Adding Dual-Memory raises the average score from 0.4521 to 0.4762 ($+5.3\%$), while adding the orchestrator (without experience memory) raises it to 0.4962 ($+9.8\%$). 
When all three components are jointly enabled, the average score further reaches 0.5107 ($+13.0\%$ over Backbone), exceeding either single addition and indicating that the two modules are complementary rather than redundant.

Regarding single-stage effects, the Dual-Memory variant shows the largest gain on Specialty Referral accuracy ($+11.5\%$), consistent with improved cross-agent state synchronization through structured memory. 
The orchestrator variant shows the largest gain on the Test Ordering ($+20.1\%$), suggesting that backtracking-driven verification refines evidence acquisition by adding missing critical tests while pruning redundant ones. 
% The full system additionally translates these upstream gains into improved Treatment IoU ($+11.5\%$).
% , indicating that better evidence collection propagates into more aligned downstream plans.

% Overall, these results demonstrate that \textbf{Dual-Memory} and the \textbf{Clinical Orchestrator} each provide measurable and complementary benefits, and their combination is crucial for achieving the full performance of ClinicalAgents.

\begin{table}[t]
\centering
 \small
\setlength{\tabcolsep}{1pt}
\renewcommand{\arraystretch}{1}
\caption{Ablation experiments, where SP denotes Specialty Referral, TO denotes Test Ordering, EX denotes Examination, Diag denotes Diagnosis, Treat denotes Treatment. The backbone model is GPT-5.2.}
\begin{tabular}{lccccccc}
\toprule
\multirow{2}{*}{Ablation} &
\multicolumn{2}{c}{SP} &
TO &
EX &
Diag &
Treat &
\multirow{2}{*}{Average} \\
\cmidrule(lr){2-3}
& Acc& IoU & IoU & DocLens & LLM-judge & IoU & \\
\midrule
Backbone & 0.5467 & 0.3045 & 0.3870 & 0.4276 & 0.5754 & 0.4716 & 0.4521 \\
+ Dual-Memory       & 0.6095 & 0.3187 & 0.4157 & 0.4460 & 0.5811 & 0.4863 & 0.4762 \\
+ Orchestrator      & 0.6125 & 0.3391 & 0.4649 & 0.4673 & 0.5914 & 0.5022 & 0.4962 \\
\midrule
\textbf{ALL}        & \textbf{0.6243} & \textbf{0.3543} & \textbf{0.4820} & \textbf{0.4800} & \textbf{0.5976} & \textbf{0.5259} & \textbf{0.5107} \\
\bottomrule
\end{tabular}%
\label{tab:ablation_medchain}
\end{table}

\subsection{Backbone Sensitivity (RQ3)} \label{sec:parameter}
In this subsection, we analyze the sensitivity of ClinicalAgents to different backbone LLMs (\textbf{RQ3}).
The results shown in Table~\ref{tab:param_sensitivity_medchain} demonstrate that ClinicalAgents is robust to backbone choice.

\begin{table}[t]
\centering
 \small
\setlength{\tabcolsep}{1pt}
\renewcommand{\arraystretch}{1}
\caption{Backbone sensitivity results. We test ClinicalAgents with different backbone models and report the performance. The symbol `—' indicates that the model is a text-only LLM and not applicable to a visual examination task.}
\begin{tabular}{@{}lccccccc@{}}
\toprule
\multirow{2}{*}{Backbone} &
\multicolumn{2}{c}
{SP} &
TO &
EX &
Diag &
Treat &
\multirow{2}{*}{Average} \\
\cmidrule(lr){2-3}
& Acc& IoU & IoU & DocLens & LLM-judge & IoU & \\
\midrule
Qwen3-VL & 0.6273 & \textbf{0.3779} & 0.3717 & 0.4369 & 0.5908 & 0.5229 & 0.4879 \\
HuaTuoGPT & 0.4948 & 0.2839 & 0.3034 & \textemdash & 0.5746 & 0.5082 & 0.4330 \\
DeepSeek-R1     & \textbf{0.6322} & 0.3621 & 0.4561 & \textemdash & 0.5427 & 0.4695 & 0.4925 \\
GPT-5.2         & 0.6243 & 0.3543 & \textbf{0.4820} & \textbf{0.4800} & \textbf{0.5976} & \textbf{0.5259} & \textbf{0.5107} \\
\bottomrule
\end{tabular}%
% \vspace{-5px}
\label{tab:param_sensitivity_medchain}
\end{table}

We use four different backbone LLMs (HuaTuoGPT-o1-7B, Qwen3-VL-235B, DeepSeek-R1, and GPT-5.2) and compare against their corresponding single-LLM baselines under the same evaluation protocol.
Overall, ClinicalAgents is moderately sensitive to the backbone, with absolute performance (Average) ranging from 0.4330 to 0.5107; yet it remains robust, consistently improving over the single-LLM counterpart across all backbones.
Specifically, ClinicalAgents increases from 0.3454 to 0.4330 for HuaTuoGPT-o1-7B (+0.0876, +25.4\%), from 0.4487 to 0.4879 for Qwen3-VL-235B (+0.0392, +8.7\%), from 0.4321 to 0.4925 for DeepSeek-R1 (+0.0604, +14.0\%), and from 0.4521 to 0.5107 for GPT-5.2 (+0.0586, +13.0\%).
Notably, the largest relative gain is observed on the weakest backbone (HuaTuoGPT-o1-7B), suggesting that ClinicalAgents provides stronger compensation when the backbone alone is less capable.
Moreover, the performance spread across backbones is reduced after applying ClinicalAgents (range decreases from 0.1067 to 0.0777 in Average), indicating improved stability against backbone variation.
These results demonstrate that ClinicalAgents can effectively leverage different backbone LLMs, consistently enhancing clinical reasoning performance while mitigating sensitivity to backbone choice.

\subsection{Efficiency Analysis (RQ4)} \label{sec:efficiency}
\begin{table}[t]
\centering
\small
\setlength{\tabcolsep}{4pt}
\caption{Efficiency comparison across representative methods under the GPT-5.2 backbone. Latency is reported as serial wall-clock time per case. Best results are in \textbf{bold}.}
\label{tab:efficiency}
\begin{tabular}{lcccc}
\toprule
\textbf{Method} & \textbf{Avg.\ Score} & \textbf{Tokens} & \textbf{Latency (s)} & \textbf{Cost (USD)} \\
\midrule
Backbone + SC-3       & 0.4573 & 45{,}265 & 731.51 & 0.1215 \\
Backbone + SC-5       & 0.4579 & 76{,}058 & 1243.62 & 0.2148 \\
Backbone + SC-7       & 0.4583 & 104{,}733 & 1681.43 & 0.2962 \\
MedAgents             & 0.4547 & 44{,}599 & 581.07 & 0.1264 \\
MedChain-Agents       & 0.4880 & 30{,}892 & 433.28 & 0.0879 \\
\textbf{ClinicalAgents} & \textbf{0.5107} & \textbf{23{,}438} & \textbf{369.07} & \textbf{0.0658} \\
\bottomrule
% \vspace{-5px}
\end{tabular}
\end{table}

In this subsection, we examine whether the gains of ClinicalAgents arise from its multi-agent orchestration or merely from additional test-time compute (\textbf{RQ4}).
% A multi-agent framework inherently consumes more inference compute than a single LLM call, raising a natural question of \emph{compute fairness}: would simply spending the same extra compute on a single model close the gap?

To answer this, we compare ClinicalAgents under the same GPT-5.2 backbone against two reference points: (i) Self-Consistency (SC-$k$), which executes the sequential backbone pipeline $k$ times independently and aggregates outputs by majority voting; and (ii) representative multi-agent baselines, MedAgents and MedChain-Agents.
For each method, we report the average score, total tokens per case, end-to-end serial latency, and API cost in Table~\ref{tab:efficiency}. 
We can summarize the key observations as follows:
\begin{itemize}
    \item \textbf{Simply scaling test-time compute yields marginal returns.} Increasing SC-$k$ from $3$ to $7$ raises the average score by only $+0.0010$ at $2.3\times$ tokens and $2.4\times$ cost.
    \item \textbf{ClinicalAgents achieves the highest score and the lowest token cost.} It attains the highest average score (0.5107) with the fewest tokens (23{,}438) and the lowest cost (0.0658 USD); against SC-7, it gains $+0.0524$ ($+11.4\%$) using only ${\sim}22\%$ of SC-7's tokens and cost. 
	% The MCTS rollouts are performed by a locally deployed, lightweight model with early stopping, which limits additional API token overhead.
    % \item \textbf{How compute is allocated matters more than how much.} Under a comparable token budget, ClinicalAgents outperforms MedAgents by $+0.0560$ while using fewer tokens ($23{,}438$ vs.\ $44{,}599$).
\end{itemize}
These results show that the gains of ClinicalAgents come from more effective compute allocation through orchestration and evidence-driven revision, rather than larger compute budgets alone.

% \clearpage

\section{Conclusions} \label{sec:conclusion}
We introduced ClinicalAgents, a multi-agent framework designed to emulate professional clinicians’ iterative, hypothesis-driven reasoning. 
ClinicalAgents departs from static, pre-scripted workflows by using an MCTS-based orchestrator that adaptively selects and reorders agent actions, enabling evidence-triggered backtracking when verification indicates missing or insufficient information. 
We further proposed a dual-memory (working memory and experience memory) design that stabilizes multi-round reasoning, enabling consistent state tracking and evidence-aware updates across iterations.
Across a five-stage workflow benchmark, ClinicalAgents achieves consistent improvements over strong single-agent and multi-agent baselines, and ablation studies validate that orchestration and dual-memory contribute complementary gains. 
ClinicalAgents provides a practical step toward more transparent and dependable agentic systems for clinical decision support.

\section{Limitations and Ethical Considerations} \label{sec:ethics}
The responsible deployment of ClinicalAgents necessitates addressing several key ethical considerations. 
First, the system must function strictly as a supportive tool to augment, rather than replace, human clinical judgment. 
Second, to mitigate potential biases inherent in LLMs, continuous monitoring across diverse demographics is essential to prevent exacerbating healthcare disparities. 
Third, stringent adherence to data privacy regulations and anonymization protocols is paramount for maintaining trust and legal compliance. 
% While ClinicalAgents holds significant promise for enhancing clinical decision-making, careful consideration of ethical, fairness, and privacy issues is essential to ensure its responsible and equitable use in healthcare.

% \section{GenAI Disclosure}
% We used generative AI tools solely for proofreading and language polishing (e.g., correcting typos and improving clarity).

\begin{acks}
This work was partially funded by the Guangdong Basic and Applied Basic Research Foundation (Project No. 2026A1515010227),
PolyU TDG25-28/TBP/11-IICA,
NSFC/RGC Joint Research Scheme (N\_PolyU5179/25), Hong Kong RGC (PolyU25600624), Innovation Technology Fund (ITS/052/23MX, PRP/009/22FX), industrial sponsors and RMGS (P0045948, P0048183, P0048191, P0046453, P0060272), NSFC under the joint application scheme (T2541073).
\end{acks}

\bibliographystyle{ACM-Reference-Format}
\balance
\bibliography{References}

% \clearpage

% \onecolumn
\appendix

\section{Supplemental Experiments and Analysis} \label{exp:qa}
In this appendix, we present a study to evaluate the generalization performance of ClinicalAgents across general medical datasets, addressing the question: \textit{How well does ClinicalAgents generalize to standard medical dialogue and QA benchmarks beyond MedChain?}

We first examine performance on four standard medical QA benchmarks, then extend the evaluation to a more challenging multi-turn clinical dialogue setting (MedDG), thereby characterizing where the gains of our framework arise and how they scale with task complexity.

\paragraph{Medical dialogue and QA datasets.}
To validate effectiveness beyond MedChain, we also evaluate on four medical QA datasets and one medical dialogue dataset:
\begin{itemize}
  \item \textbf{MedQA}~\cite{jin2021disease}: we report results on the official \textbf{US test} subset.
  \item \textbf{PubMedQA}~\cite{jin2019pubmedqa}: we report results on its official \textbf{test split} (PQA-L), with answers in \{yes, no, maybe\}.
  \item \textbf{MedBullets}~\cite{chen2025benchmarking}: we report results on the official \textbf{test split} for multiple-choice clinical QA.
  \item \textbf{PathVQA}~\cite{he2020pathvqa}: we filter the evaluation to \textbf{yes/no-type questions} and cast them into a standardized binary-choice format for comparability.
  \item \textbf{MedDG}~\cite{liu2022meddg}: an entity-centric multi-turn medical consultation dataset for entity-aware medical dialogue generation; we report BLEU, ROUGE, and Entity-F1 on its official test split.
\end{itemize}

% For all four QA datasets, we report Accuracy following standard practice.
% For MedDG, we report BLEU-1, BLEU-4, ROUGE-1, ROUGE-2, and Entity-F1, following the official evaluation protocol.

\paragraph{Results on medical QA}
As shown in Table~\ref{tab:generalization}, ClinicalAgents generalizes well to four standard open-domain medical QA benchmarks and achieves the best overall Average of 0.8053. 
It outperforms the strongest multi-agent baseline MedChain-Agents (0.7990) by +0.0063 (+0.8\%), indicating that our framework remains beneficial even on single-turn QA tasks, not just on a multi-stage pipeline. 
The gains are most evident on PubMedQA (0.7660) and MedBullets (0.8279), suggesting improved answer selection via better information consolidation.
In contrast, the gap on MedQA is marginal (0.8995 vs.\ 0.9002), where strong backbones already perform near saturation. 
In such relatively simple QA settings, the orchestrator’s verification/backtracking loop provides less additional benefit, and the remaining gains are mainly attributed to the Dual-Memory that stabilizes intermediate reasoning states. 
Finally, performance on PathVQA (0.7278) is constrained by the backbone’s visual understanding capability, limiting further improvement from coordination alone.
We further computed 95\% CIs and paired bootstrap significance tests for all methods. 
Because static, single-turn queries rarely trigger our MCTS-based backtracking mechanism, the improvements on these QA datasets are not statistically significant ($p > 0.05$); nevertheless, ClinicalAgents still maintains the highest overall average.

\begin{table*}[!htbp]
\centering
\small
\caption{Generalization results on four open-domain medical QA datasets. For text-only models, the average is computed over applicable text-only datasets.  The best result for each metric is highlighted in bold, and the second-best result is underlined. The symbol `—' indicates that the model is a text-only LLM and not applicable to the visual QA task.}
\label{tab:generalization}
\setlength{\tabcolsep}{6pt}
\renewcommand{\arraystretch}{1}
\begin{tabular}{llccccc}
\toprule
\multirow{2}{*}{\textbf{Category}} &
\multirow{2}{*}{\textbf{Methods}} &
\textbf{MedQA} & \textbf{PubMedQA} & \textbf{MedBullets} & \textbf{PathVQA} &
\multirow{2}{*}{\textbf{Average}} \\
& & \textbf{Acc.} & \textbf{Acc.} & \textbf{Acc.} & \textbf{Acc.} & \\
\midrule

\multirow{8}{*}{\textbf{Single LLM}}
& GPT-5.2                   & 0.8845 & 0.6820 & 0.7890 & 0.6835 & 0.7598 \\
& DeepSeek-R1               & 0.8987 & 0.6200 & 0.7987 & \textemdash & 0.7725 \\
& Qwen3-VL-235B             & 0.8209 & 0.5620 & 0.6721 & \underline{0.7377} & 0.6982 \\
& Intern-S1                 & 0.8256 & 0.7580 & 0.7240 & 0.7374 & 0.7613 \\
& Llama-4-Maverick-17B      & 0.8798 & 0.6880 & 0.7695 & 0.6880 & 0.7563 \\
& MedGemma-4B               & 0.6677 & 0.6680 & 0.5812 & 0.5455 & 0.6156 \\
& HuaTuoGPT-o1-7B           & 0.7101 & 0.5520 & 0.5974 & \textemdash & 0.6198 \\
& FineMedLM-o1-8B           & 0.6787 & 0.6440 & 0.6039 & \textemdash & 0.6422 \\
\midrule

\multirow{3}{*}{\textbf{Single Agent}}
& Few-shot + CoT            & 0.8916 & 0.7460 & 0.7987 & 0.7017 & 0.7845 \\
& RAG                       & 0.8963 & 0.7180 & 0.7922 & 0.7112 & 0.7794 \\
& ReAct                     & 0.8947 & 0.7320 & 0.8019 & 0.7174 & 0.7865 \\
\midrule

\multirow{7}{*}{\textbf{Multi-Agents}}
& ReConcile~\cite{chen2024reconcile}          & 0.8924 & \underline{0.7640} & 0.8052 & 0.7124 & 0.7935 \\
& AutoGen~\cite{wu2024autogen}                & 0.8767 & 0.7140 & 0.7955 & 0.7344 & 0.7801 \\
& MedAgents~\cite{tang2024medagents}          & 0.8790 & 0.7240 & 0.8084 & 0.6829 & 0.7736 \\
& MDAgents~\cite{kim2024mdagents}             & 0.8932 & 0.7420 & 0.7727 & \textbf{0.7397} & 0.7869 \\
& ColaCare~\cite{wang2025colacare}            & 0.8932 & 0.7240 & 0.7662 & 0.6901 & 0.7684 \\
& MedChain-Agents~\cite{liu2024medchain}      & \textbf{0.9002} & 0.7420 & \underline{0.8182} & 0.7356 & \underline{0.7990} \\
& \textbf{Ours}                              & \underline{0.8995} & \textbf{0.7660} & \textbf{0.8279} & 0.7278 & \textbf{0.8053} \\
\bottomrule
\end{tabular}
\end{table*}

\paragraph{From marginal to significant: scaling to complex clinical dialogue.}
To better probe our complex reasoning capabilities, we conduct additional experiments on a more challenging dataset, MedDG~\cite{liu2022meddg}. 
As shown in Table~\ref{tab:meddg}, ClinicalAgents consistently outperforms representative baselines across all five metrics. 
It scores 42.67 on BLEU-1 and 23.36 on BLEU-4, outperforming the strongest baseline, MedChain-Agents, by 3.82\% and 5.99\%, respectively. 
For ROUGE-1, ClinicalAgents obtains 18.43, improving over MedAgents by 24.11\%. 
For ROUGE-2 and Entity-F1, ClinicalAgents achieves 8.42 and 20.72, surpassing MedChain-Agents by 27.58\% and 22.89\%, respectively. 
Furthermore, ClinicalAgents achieves statistically significant improvements ($p < 0.05$) over the strongest baseline (\underline{underlined}) on each metric.
This indicates that the orchestrator's hypothesize-then-verify loop effectively grounds the generated responses in clinically relevant entities, while experience memory retrieves precedent dialogues to stabilize entity selection across turns.
 
\begin{table}[!htbp]
\centering
\footnotesize
\renewcommand{\arraystretch}{1}
\setlength{\tabcolsep}{3pt}
\caption{Performance comparison across representative methods on MedDG. The best result for each metric is highlighted in bold, and the second-best result is underlined.}
\label{tab:meddg}
\begin{tabular}{lccccc}
\toprule
\textbf{Method} & \textbf{BLEU-1} & \textbf{BLEU-4} & \textbf{ROUGE-1} & \textbf{ROUGE-2} & \textbf{Entity-F1} \\
\midrule
GPT-5.2         & 39.88 & 21.25 & 13.14 & 4.56 & 15.59 \\
MedAgents       & 40.87 & 21.60 & \underline{14.85$\pm$0.79} & 4.63 & 16.74 \\
MedChain-Agents & \underline{41.10$\pm$1.06} & \underline{22.04$\pm$0.74} & 14.40 & \underline{6.60$\pm$0.71} & \underline{16.86$\pm$1.06} \\
\textbf{ClinicalAgents} & \textbf{42.67$\pm$0.87} & \textbf{23.36$\pm$0.62} & \textbf{18.43$\pm$0.86} & \textbf{8.42$\pm$0.63} & \textbf{20.72$\pm$1.17} \\
\bottomrule
\end{tabular}
\end{table}

\section{Case Study} \label{sec:case_study}
We present two representative cases that illustrate when the backtracking mechanism succeeds and when it fails.

\paragraph{Success case.}
Table~\ref{tab:case_study_success} shows a multi-pathology case (ground truth: \textit{Lung Cancer, Tuberculosis, Pneumonia}) where unexpected imaging findings trigger a productive revision from a single-infection hypothesis to a joint malignancy-and-infection diagnosis.

\begin{table}[htbp]
\centering
\small
\caption{Success case. Ground truth: \textit{Lung Cancer, Tuberculosis, Pneumonia}.}
\label{tab:case_study_success}
\renewcommand{\arraystretch}{0.85}
\begin{tabular}{p{0.22\linewidth} p{0.71\linewidth}}
\toprule
\textbf{Phase} & \textbf{Description} \\
\midrule
\textbf{Perceive (Initial)}
& Symptoms: recurrent cough, purulent sputum, low-grade fever, and weight loss. \\
\midrule
\textbf{Hypothesize$_1$}
& Based on symptoms, the system initially hypothesizes \textit{Typical Bacterial Pneumonia}. \\
\midrule
\textbf{Verify$_1$}
& Orders Chest X-ray, CBC, CRP. Elevated markers (CRP/neutrophils) support infection, but X-ray unexpectedly reveals a spiculated mass and apical cavitation. Unexplained structural abnormalities trigger backtracking. $\rightarrow$ \textbf{Backtrack.} \\
\midrule
\textbf{Hypothesize$_2$}
& The single-infection belief is insufficient. The system pivots to co-evaluate malignancy and chronic infection, updating its hypothesis to: \textit{Lung Cancer, Tuberculosis, and Pneumonia}. \\
\midrule
\textbf{Verify$_2$ (Final)}
& Orders targeted tests (high-resolution CT, biopsy, GeneXpert, sputum culture). Results definitively confirm Lung Cancer alongside tuberculosis and pneumonia. \\
\bottomrule
\end{tabular}
\end{table}

\paragraph{Failure case.}
Table~\ref{tab:case_study_failure} shows a multisystem case (stage~IV cervical cancer after chemo-radiotherapy) where the orchestrator triggers backtracking twice and correctly identifies missing evidence, but several ground-truth conditions remain unresolved because the required confirmatory examinations are not available in the case record.
% , which is a data-bounded rather than reasoning-bounded failure.

\begin{table}[htbp]
\centering
\small
\caption{Failure case. Ground truth: \textit{Viral Hepatitis Spectrum, Sinus Tachycardia, Pleural Effusion, Hydronephrosis, Pulmonary Infection}.}
\label{tab:case_study_failure}
\renewcommand{\arraystretch}{0.85}
\begin{tabular}{p{0.22\linewidth} p{0.71\linewidth}}
\toprule
\textbf{Phase} & \textbf{Description} \\
\midrule
\textbf{Perceive (Initial)}
& Symptoms: Stage IV cervical cancer after chemo-radiotherapy, rectal bleeding, history of HBV infection, ureteral stricture, and hydronephrosis. \\
\midrule
\textbf{Hypothesize$_1$}
& Broad hypothesis set: \textit{Sinus Tachycardia, Pleural Effusion, Hydronephrosis, Pulmonary Infection, Viral Hepatitis Spectrum, Pericardial Effusion, Radiation Proctitis}, and \textit{Acute Blood Loss Anemia}. \\
\midrule
\textbf{Verify$_1$}
& Orders ECG, chest X-ray, CBC, liver panel, renal panel. \textit{Sinus Tachycardia} and \textit{Pleural Effusion} are supported, and \textit{Acute Blood Loss Anemia} is excluded by CBC, but others remain unconfirmed. $\rightarrow$ \textbf{Backtrack.} \\
\midrule
\textbf{Hypothesize$_2$}
& Updates hypothesis set: \textit{Pulmonary Infection}, \textit{Hydronephrosis}, \textit{Viral Hepatitis Spectrum}, \textit{Pericardial Effusion}, and \textit{Radiation Proctitis} remain active. \\
\midrule
\textbf{Verify$_2$}
& Orders chest CT, abdominal ultrasound, HBV-DNA, hepatitis serology panel. \textit{Pulmonary Infection} is strengthened, but others remain unresolved due to partial evidence return. $\rightarrow$ \textbf{Backtrack.} \\
\midrule
\textbf{Hypothesize$_3$}
& Updates hypothesis set: \textit{Hydronephrosis} and \textit{Viral Hepatitis Spectrum} remain unresolved. \\
\midrule
\textbf{Verify$_3$ (Final)}
& Orders pelvic MRI and contrast-enhanced abdominal CT. Confirmatory evidence is unavailable. \textbf{Final outcome:} confirmed \textit{Sinus Tachycardia, Pleural Effusion, Pulmonary Infection}; excluded \textit{Acute Blood Loss Anemia}; unresolved \textit{Hydronephrosis, Viral Hepatitis Spectrum, Pericardial Effusion, Radiation Proctitis} due to missing follow-up evidence. \\
\bottomrule
\end{tabular}
\end{table}

\section{Prompt Templates} \label{prompts:agents}
Here we provide the prompt for LLM-as-a-Judge used in our experiments (task 4) in the MedChain benchmark.
Other prompt templates for the doctor agents are provided in the repository of our code.

\begin{promptbox}{Diagnosis correctness evaluation prompt}
You are a physician specializing in diagnostic evaluation. Below, a diagnosis result will be provided alongside the ground-truth diagnosis. Your task is to assess the accuracy of the diagnosis result by comparing it to the ground truth. Use the following criteria to assign an accuracy score from 1 to 5. You should only output the score as a single number without any explanation.

Evaluation Logic (Step-by-Step):
1. Normalization: Treat synonyms as identical (e.g., "URI" == "Upper Respiratory Infection", "Ca" == "Cancer", "HBP" == "Hypertension").
2. Core Match: Does the prediction capture the primary pathology?
   - GT: "Appendicitis" vs Pred: "Acute Appendicitis" -> Score 5 (Match).
   - GT: "Acute Appendicitis" vs Pred: "Gastritis" -> Score 1 (Mismatch).
3. Granularity:
   - If GT is specific ("Type 2 Diabetes") and Pred is general ("Diabetes"), deduct 1 point (Score 4).
   - If GT is general ("Fracture") and Pred is specific ("Tibial Fracture"), accept as correct (Score 5).

Scoring Rubric:
- 1 (Wrong): The diagnosis is completely inaccurate, with no elements matching the ground truth diagnosis. e.g., appendicitis is diagnosed as a migraine.
- 2 (Major Miss): The diagnosis is mostly inaccurate, with only a few elements matching the ground truth diagnosis, and the overall deviation is significant. e.g., asthma is diagnosed as pneumonia.
- 3 (Partial): The diagnosis is partially accurate, with some correct elements that are relevant to the ground truth diagnosis, but the overall result shows large discrepancies. e.g., Crohn's disease is diagnosed as irritable bowel syndrome (IBS).
- 4 (Minor Issue): The diagnosis is mostly accurate, with most elements matching the ground truth diagnosis, but there are minor errors or omissions. e.g., type 1 diabetes is diagnosed as type 2 diabetes.
- 5 (Perfect): The diagnosis is completely accurate and matches the ground truth diagnosis entirely, or it is a different name for the same condition, with no significant difference in meaning. e.g., "myocardial infarction" vs. "heart attack", or "upper respiratory infection" vs. "URI".
\end{promptbox}

\clearpage

\end{document}